\documentclass{article}

\usepackage[final, nonatbib]{neurips_2020}

\usepackage[utf8]{inputenc} % allow utf-8 input
\usepackage[T1]{fontenc}    % use 8-bit T1 fonts
\usepackage{hyperref}       % hyperlinks
\usepackage{url}            % simple URL typesetting
\usepackage{booktabs}       % professional-quality tables
\usepackage{amsfonts}       % blackboard math symbols
\usepackage{nicefrac}       % compact symbols for 1/2, etc.
\usepackage{microtype}      % microtypography

\usepackage{subfiles}
\usepackage{tabularx}
\usepackage{graphicx}
\usepackage{lipsum}
\usepackage{enumitem}

\newcommand\blfootnote[1]{%
  \begingroup
  \renewcommand\thefootnote{}\footnote{#1}%
  \addtocounter{footnote}{-1}%
  \endgroup
}
\usepackage{amsmath}
\DeclareMathOperator*{\argmax}{arg\,max}

\title{Learning Task-agnostic Representation via Toddler-inspired Learning }

\author{
Kwanyoung Park\qquad
Junseok Park\qquad
Hyunseok Oh\qquad
Byoung-Tak Zhang*\qquad
Youngki Lee*
\\
Department of Computer Science and Engineering, AI Institute (AIIS)\\
Seoul National University, Seoul 08826, South Korea\\
{\tt\small \{william202, jspark227, ohsai\}@snu.ac.kr, 
btzhang@bi.snu.ac.kr, youngkilee@snu.ac.kr}
}

\begin{document}

\maketitle
%%%%%%%%% ABSTRACT
\begin{abstract}

One of the inherent limitations of current AI systems, stemming from the passive learning mechanisms (e.g., supervised learning), is that they perform well on labeled datasets but cannot deduce knowledge on their own. 
To tackle this problem, we derive inspiration from a highly intentional learning system via action: the toddler. 
Inspired by the toddler's learning procedure, we design an interactive agent that can learn and store task-agnostic visual representation while exploring and interacting with objects in the virtual environment.
Experimental results show that such obtained representation was expandable to various vision tasks such as image classification, object localization, and distance estimation tasks. In specific, the proposed model achieved
100\%, 75.1\% accuracy and 1.62\% relative error, respectively, which is noticeably better than autoencoder-based model (99.7\%, 66.1\%, 1.95\%), and also comparable with those of supervised models (100\%, 87.3\%, 0.71\%).

\end{abstract}

\blfootnote{*Corresponding authors.}

%%%%%%%%% BODY TEXT
\section{Introduction}

Although recent deep learning methods are showing an overwhelming performance in the computer vision domain~\cite{voulodimos2018cvsurvey}, there are major limitations of these data-driven learning:
(i) a well-distributed large labeled dataset is needed to learn a feature properly~\cite{DBLP:journals/corr/abs-1811-06052}, (ii) they are task-specific in the sense that adapting to multiple different tasks or transfer learning is difficult~\cite{lake2017buildingmachines}.

Several learning frameworks are suggested to overcome each limitation, but none of them are fit to solve both of them due to their drawbacks. 
Semi-supervised learning trains the model using both the labeled data and extra unlabeled data to reduce the labeling cost.
However, it increases the sample complexity only by a constant factor compared to supervised learning without strong assumptions on unlabeled data~\cite{lu2009semisupervisedlearning}.
Meta-learning trains in a set of well-known tasks and leverage the acquired knowledge in learning a similar task. 
Still, there should be a sufficient number of well-labeled prior tasks that are much similar to the target task~\cite{kang2018transferablemetalearning}.
Multi-task learning effectively adapts to a set of related tasks simultaneously by forming an inductive bias from intrinsic dependencies of tasks, but it is challenging to optimize shared parameters in each task's competing objectives~\cite{liu2019multitasklearningauxilarytask}.

These challenges stem from the difference in how human and data-driven AI models the learning process~\cite{lake2017buildingmachines}. Data-driven AI takes a statistical pattern recognition-based approach, so the knowledge accumulation relies on the observed data. 
In contrast, humans actively inspect the environment to collect data and learn with only a few of them by building a generalized world model. Thus, a solution could be to integrate the learning process of human into AI techniques. 

ML researchers are recently taking an interest in learning of a child to seek technical advances~\cite{bambach2018toddlerinspiredvisualobject,schank1972nlp,turing2009child}.
Learning properly in a child stage is crucial for learning-to-learn capabilities like goal-setting, self-control ~\cite{zosh2017learnthruplayreview}. 
Life experiences and prior knowledge learned in childhood are known to influence the learning of a grown-up.
The latest works in deep learning like visual object learning~\cite{bambach2018toddlerinspiredvisualobject} 
are seeking advance from how children learn. By understanding how the child learns, we can understand how learning-to-learn capabilities and task-agnostic knowledge of objects are nurtured.

In this work, we propose a new learning framework to deal with the challenges, inspired by a highly intentional learning system via action: the toddler.
Toddlers unconsciously learn through interaction and play with their surrounding environment, rather than self-directed task-specific learning of an adult~\cite{mcdonough2013difference_adult_child}. Large-scale studies suggest that general understandings of objects develop in an early stage of the toddler without any supervision, through interaction on objects like mouthing, chewing, and rotating~\cite{gibson1988toddlerlearning,piaget1952origins}.
Furthermore, teaching formal subjects too early for a child is counterproductive in the long run~\cite{suggate2013childlatelearning}, and children learn cognitive or self-regulatory abilities through playing~\cite{piaget1952origins,zosh2017learnthruplayreview}.
These studies motivated us to organize unsupervised or weakly-supervised learning through play in an interactive and playful environment to simulate how toddlers learn.

We formulate a toddler-inspired learning framework and simulate the exploration and interaction-based learning in an interactive environment. In particular, we first designed a virtual environment where the agent can freely roam and interact with objects and gets feedback (reward). Second, we designed the agent's network architecture to extract the visual knowledge of objects to the embedding called \emph{interaction feature maps}. Interaction feature maps are designed to have only one feature image per interaction to make the agent learn a compact interaction-based representation.
Finally, we transferred the visual knowledge to downstream computer vision tasks by using the interaction feature map as prior.
Learning downstream tasks with the interaction feature map were able to achieve 99.7\%, 62.8\% accuracy, and 3.0\% relative error in image classification, object localization task, and distance estimation, which is 0.3\%, 16.9\%, 13.6\% better than the autoencoder-based unsupervised transfer learning. Moreover, the number of images to develop the embedding prior was notably smaller than the unsupervised counterpart. It shows that the toddler-inspired learning framework can efficiently gain a transferrable knowledge of objects with active interaction-based data collection.
%\subfile{Sections/2_Background}
\section{Methods}
\textbf{Interactive Virtual Environment.}
Motivated by~\cite{gibson1988toddlerlearning}, we designed an environment supporting the human-like visual observation and active physical interaction with the object, to train the visual knowledge prior without any explicit labels.  We used VECA~\cite{VECA}, a virtual environment generation toolkit for human-like agents, to implement the environment. The environment's reward structure provides a sparse positive reward signal when the agent is touching or playing with the prop objects, and provides a near-zero negative reward to aid the navigation to the prop object. This reward structure motivates the agent to visually locate distant objects while freely exploring and observing objects in depth when it is nearby. The agent collects data without any label and establishes a more profound visual understanding of an object compared to the unsupervised learning on object images without any context.

\textbf{Toddler-inspired learning.} 
With the interactive environment, we formulate the toddler-inspired learning framework, which aims to acquire a general understanding of objects without any supervision, but by exploring the environment and interacting with the objects, as a toddler does. 
We assume that the agent can only visually observe and interact with the object during the sparsely rewarded reinforcement learning task. With the reinforcement learning, we want the agent to learn a transferable representation embedding $f_{\theta}(x)$ parameterized by $\theta$ through sufficient observation and interaction, without any supervision of the downstream tasks. The representation mapping $f_{\theta}(x)$ will be a general prior for supervised downstream tasks $\mathbb{T} = \{T_i\}_{i = 1,\cdots,n}$ with datasets $\mathcal{D}_{\mathbb{T}} =\{\mathcal{D}_{T_{i}}\}$
%,$ D_{T_{i}} = {(x_j, y_j)}$ %\{(\mathcal{X}_{T_{i}}, \mathcal{Y}_{T_{i}})\}$
%_{i = 1,\cdots,n}
collected from the environment. 

\begin{figure*}[t]
    \centering
    \includegraphics[width=\textwidth]{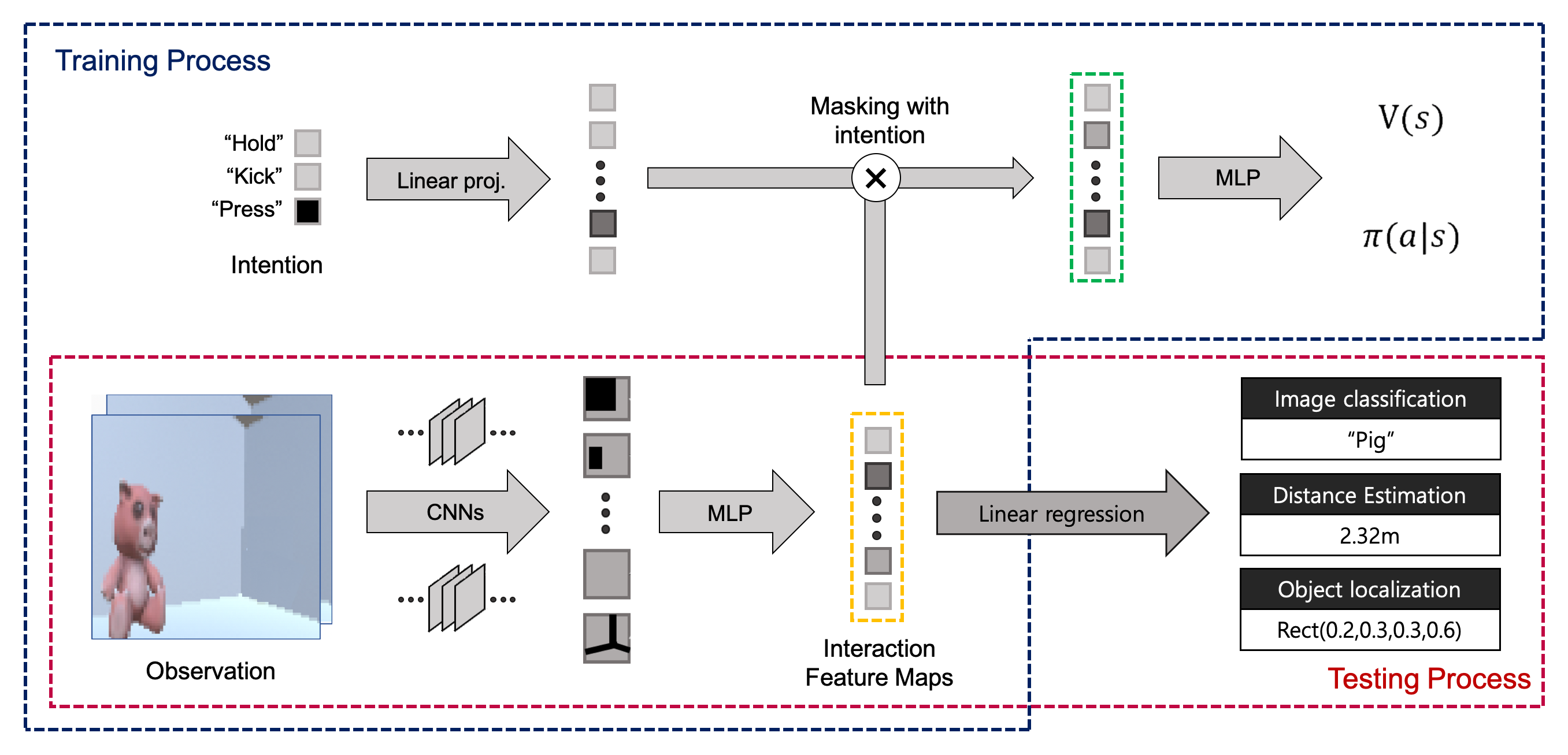}
    \caption{Overview of  toddler-inspired learning framework and network architecture.}
    \vspace{-10px}
    \label{figure:Overview}
\end{figure*}

The framework consists of two phases. First, the transferable representation $f_{\theta}$ is pretrained in an interactable environment, by solving a reinforcement learning problem in Eq. 1. $o_t$, $a_t$ represents observation, action in time $t$, while $r(\cdot)$, $\pi_{\psi}(\cdot | \cdot)$ and $\gamma$ are reward function, policy and discount factor, respectively. Second, we use the representation mapping $f_{\theta}(x)$ as an embedding prior and transfer to the downstream tasks $\mathbb{T}$. We evaluate the generality of representation with its transferability, which results in maximizing the sum of each objective shown in Eq. 2. and $\{J_{T_i}\}$ denotes the set of the objective function for each task $\{T_{i}\}$. Please note that we cannot directly optimize the formula since the task distribution is unknown while training.
\begin{align}
%\textrm{Maximize} ~~ J_{train}(f(\cdot)) = \sum_{t=0}^{\infty} {\gamma}^t r(s_t, \pi_{\psi}(a_t|f_{\theta}(o_t))) \\
\hat{\theta}, \hat{\psi} = \argmax_{\theta, \psi}J_{train}(f_{\theta}, \pi_{\psi}) = \argmax_{\theta, \psi} \sum_{t=0}^{\infty} {\gamma}^t r(s_t, \pi_{\psi}(a_t|f_{\theta}(o_t))) \\
\textrm{to indirectly maximize} ~~ J_{test}(z) 
= \sum_{T_i \in \mathbb{T}} \sum_{(x,y) \in \mathcal{D}_{T_{i}}} J_{T_i}(f_{\hat{\theta}}(x), y)
\end{align}
The Fig. \ref{figure:Overview}. shows the overview of the procedure in our toddler-inspired learning framework. In the first phase, the agent trains under the reinforcement learning framework and learns the efficient transferable representation, which we named as \emph{interaction feature maps}, through exploring the environment and interacting with the object. In the second phase, the representation embedding becomes a feature extractor of the data points on downstream tasks. In this figure, the representation embedding's generality is evaluated with the transferability to three vision tasks: image classification, distance estimation, and object localization.

\textbf{Network Architecture.} To learn and store transferable knowledge, we designed the architecture of the agent as Fig.~\ref{figure:Overview}. Visual observation of the agent is encoded with CNN and MLP, resulting in \emph{interaction feature maps}. 
Those feature maps are masked with linearly embedded intention and determine the action of the agent.
Since the agent's movement only depends on masked features, the agent must learn to represent abstract features of the object corresponding to its interaction.
\section{Experiments}

\subsection{Experimental Setup}
%To evaluate the generality of knowledge gained during exploration, we measure the transfer performance with other vision tasks.
%To evaluate our proposed learning procedure, we measured the generality of knowledge gained during exploration with its transfer performance.
To show that the agent could acquire transferable knowledge through the toddler-inspired learning framework, we evaluated the interaction feature map’s transfer performance on three supervised visual downstream tasks: image classification, distance estimation, and object localization. In specific, we fixed the parameters of the interaction feature map $f_\theta$ after pretraining and connected linear layers for transfer learning. 
We used VECA~\cite{VECA} toolkit to implement an interactive 3D environment that includes three prop objects (toy pyramid, ball, doll) and a baby agent. The agent receives RGB 84x84 binocular vision and has two kinds of action - movement and interaction(hold, kick, press).
%The movement of the agent is represented as a two-dimensional vector (walking velocity). The agent is rewarded when the agent successfully navigated to the object and interact with the object. 
The reward signal differs for each object-interaction pair. It could be positive (If the baby presses the doll, then the doll will make sound and the baby will be joyful) or negative (If the baby kicks the toy pyramid, then the baby will feel pain). 

%\textbf{Dataset.} We collected 2400 image data by randomly rotating and positioning the objects used in the environment. For all experiments, we split the data into 2100/300 images and used them for training/testing. 

%Performance of the transferred agent reflects its learning procedure, but also affected by its architecture.
\textbf{Baselines.} Since the performance of the agent is related to its architecture, we compare the performance of the agent to baseline agents with the same architecture but different training methods:
\raggedbottom
\begin{itemize}[noitemsep,topsep=0pt,wide,labelindent=0cm,leftmargin=0.35cm]

\item Random: Randomly initialize the agent's network parameters and only train the connected layer. It shows the naive performance from the architecture.

\item Autoencoder: The agent's network is trained as an autoencoder. We use the performance as a baseline of representation learning without explicitly labeled data.

\item Supervised: Both agent's network and the connected network are trained supervised (without transferring) for a specific downstream task. We interpret its performance as an optimal achievable performance with this architecture.

\end{itemize}

\subsection{Results and Discussion}

\begin{figure*}[t]
    \centering
    \includegraphics[width=\textwidth]{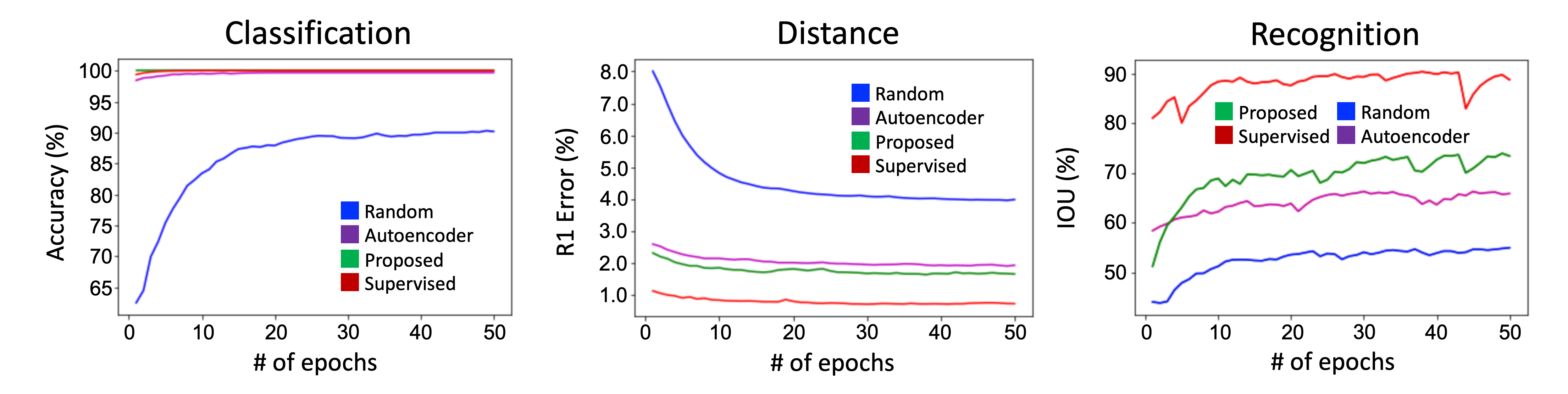}
    \caption{Learning curve of transfer learning. For classification and recognition, higher is better. For distance estimation, lower is better. Best viewed in color.}
    \label{figure:Agent}
\end{figure*}

\iffalse
\begin{table}[h]
\begin{tabular}{|>{\centering\arraybackslash}m{0.4\textwidth}|>{\centering\arraybackslash}m{0.11\textwidth}|>{\centering\arraybackslash}m{0.11\textwidth}|>{\centering\arraybackslash}m{0.11\textwidth}|>{\centering\arraybackslash}m{0.11\textwidth}|}
\hline
Task (metric)                           & Random & Autoencoder & Proposed & Supervised \\ \hline
Classification (top-1 accuracy)         & 90.0\% & 99.7\%      & 100\%    & 100\%  \\ \hline
Distance estimation (relative L1 error) & 4.01\% & 1.95\%      & 1.62\%   & 0.71\%  \\ \hline
Recognition (intersection of unions)    & 55.2\% & 66.1\%      & 75.1\%   & 87.3\%  \\ \hline
\end{tabular}
\vspace{5px}
\caption{Numerical transfer performance of various models. For classification and recognition, higher is better. For distance estimation, lower is better.}
\label{table:Results}
\end{table}
\fi

\begin{table}[h]
\begin{tabular}{|>{\centering\arraybackslash}m{0.32\textwidth}|>{\centering\arraybackslash}m{0.13\textwidth}|>{\centering\arraybackslash}m{0.13\textwidth}|>{\centering\arraybackslash}m{0.13\textwidth}|>{\centering\arraybackslash}m{0.13\textwidth}|}
\hline
Task (Metric, \%)                  & Random & Autoencoder & Proposed & Supervised \\ \hline
Classification (Accuracy)          & 90.0$\pm$2.9  & 99.7$\pm$0.0  & 100$\pm$0     & 100$\pm$0 \\ \hline
Distance estimation (L1 error)     & 4.01$\pm$0.61 & 1.95$\pm$0.08 & 1.62$\pm$0.02 & 0.71$\pm$0.01 \\ \hline
Recognition (IOU)                  & 55.2$\pm$1.2  & 66.1$\pm$0.5  & 75.1$\pm$1.4  & 87.3$\pm$0.8 \\ \hline
\end{tabular}
\vspace{5px}
\caption{Numerical transfer performance with standard errors. For classification and recognition, higher is better. For distance estimation, lower is better.}
\label{table:Results}
\end{table}

As shown in Table \ref{table:Results} and Fig. \ref{figure:Agent}, the transferred models achieved better performance than random and autoencoder models, while being also comparable to supervised models. In specific, the proposed model was able to achieve relative improvement of 0.3\%, 16.9\%, 13.6\% for classification, distance estimation, recognition, from autoencoder models. It shows that the agent was able to learn task-agnostic knowledge from the environment without any explicit labels.

Why would the agent learn those features? For the classification and recognition, we suppose that it is because the agent must recognize and classify the objects to achieve maximal reward, while action of the agent is dependent on the transferred feature. 
For the distance estimation, we suppose that this ability comes from the training of the critic (value function). Since the critic has to predict the cumulative reward, the agent would learn how much time will be required to reach the object, thus roughly estimate the distance to the object.
\section{Conclusion \& Future Works}

Inspired by how toddlers learn, we proposed a toddler-inspired learning framework to gain transferable visual knowledge of objects by exploring and interacting with the environment. We evaluated its transfer performance to several supervised downstream visionary tasks. Evaluation results show that the agent could gain a transferable knowledge of objects by exploring and interacting with the environment. 

However, our method is still far from how a toddler learns. We used hand-crafted reward and applied a conventional reinforcement learning algorithm to train the agent. We suggest that substituting hand-crafted reward for intrinsic reward and developing a human-like fast and adaptive learning algorithm would be a promising future direction.

\section*{Acknowledgement}

This work was supported by Institute for Information \& Communications Technology Planning \& Evaluation(IITP) grant funded by the Korea government(MSIT) (No.2019-0-01367, Infant-Mimic Neurocognitive Developmental Machine Learning from Interaction Experience with Real World (BabyMind)).

{\small
\bibliographystyle{nips}
\bibliography{egbib}
}

\section{Technical appendix}

\subsection{Detailed description of training process} At the start of the each episode, the agent has an intention to do certain interaction (e.g. in this episode, the baby wants to kick something). At the initial point, the agent randomly explores and interacts with the objects, receiving both positive and negative rewards. During the training process, the agent learns to find the object matching its intention to maximize the reward. 

\subsection{Dataset}
We collected 2400 binocular RGB image data by randomly rotating and positioning the objects used in the environment. For all experiments, we split the data into 2100/300 images and used them for training/testing. 

\subsection{Technical details}

\textbf{Architecture.} We used the CNN architecture introduced in \cite{DQN}, and connected two layers to make 512-dimensional interaction feature maps. Those features are transfered with single linear layer. 

\textbf{Image classification.} Like casual image classification tasks, the agent has to classify images by its included object. We used cross-entropy loss with softmax activation to train the model.

\textbf{Distance estimation.} The agent has to estimate the distance between the camera and the object. The distance is log-normalized to have zero mean and unit variance. We used mean squared error loss to train the model.

\textbf{Object localization.} The agent has to localize the object with a bounding box. Coordinate of vertices of bounding box is within the range of (0, 1). We designed the network to output four values: output center coordinate, width and height of the bounding box. 

\textbf{Training.} While training in the virtual environment, we used Adam\cite{Adam} optimizer with learning rate of $0.00025$ and trained the agent using SAC algorithm\cite{haarnoja2018soft} for 3.2M frames. For transferring the agent, we used the same optimizer with learning rate of $0.001$ and trained the agent for 50 epochs. 

\iffalse
\begin{table}[h]
\begin{tabular}{|>{\centering\arraybackslash}m{0.24\textwidth}|>{\centering\arraybackslash}m{0.15\textwidth}|>{\centering\arraybackslash}m{0.15\textwidth}|>{\centering\arraybackslash}m{0.15\textwidth}|>{\centering\arraybackslash}m{0.15\textwidth}|}
\hline
Task                  & Random & Autoencoder & Proposed & Supervised \\ \hline
Classification        & 90.0$\pm$2.9  & 99.7$\pm$0.0  & 100$\pm$0     & 100$\pm$0 \\ \hline
Distance estimation   & 4.01$\pm$0.61 & 1.95$\pm$0.08 & 1.62$\pm$0.02 & 0.71$\pm$0.01 \\ \hline
Recognition           & 55.2$\pm$1.2  & 66.1$\pm$0.5  & 75.1$\pm$1.4  & 87.3$\pm$0.8 \\ \hline
\end{tabular}
\vspace{5px}
\caption{Numerical transfer performance with standard errors. For classification and recognition, bigger is better. For distance estimation, smaller is better.}
\label{table:Results}
\end{table}
\fi

\end{document}